# Modeling and Application of Series Elastic Actuators for Force Control Multi Legged Robots

Arumugom.S, Muthuraman.S, Ponselvan.V

**Abstract**— Series Elastic Actuators provide many benefits in force control of robots in unconstrained environments. These benefits include high force fidelity, extremely low impedance, low friction, and good force control bandwidth. Series Elastic Actuators employ a novel mechanical design architecture which goes against the common machine design principal of "stiffer is better". A compliant element is placed between the gear train and driven load to intentionally reduce the stiffness of the actuator. A position sensor measures the deflection, and the force output is accurately calculated using Hooke's Law (F=Kx). A control loop then servos the actuator to the desired output force. The resulting actuator has inherent shock tolerance, high force fidelity and extremely low impedance. These characteristics are desirable in many applications including legged robots, exoskeletons for human performance amplification, robotic arms, haptic interfaces, and adaptive suspensions. We describe several variations of Series Elastic Actuators that have been developed using both electric and hydraulic components.

**Index Terms**— Series Elastic Actuators, legged robots, force control.

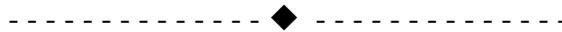

## 1. INTRODUCTION

In traditional manufacturing operations, robots perform tedious and repetitive tasks with great speed and precision. In this setting, where the environment is controlled and the tasks are repetitive, position controlled robots which trace predefine joint trajectories are optimal. However, in highly unstructured environments, where little is known of the environment, force controlled robots that can comply to the surroundings are desirable. This is the case for legged robots walking over rough terrain, robotic arms interacting with people, wearable performance-enhancing exoskeletons, haptic interfaces, and other robotic applications.

An ideal force-controllable actuator would be a perfect force source, outputting exactly the commanded force independent of load movement. In the real world, all force-controllable actuators will have limitations that result in deviations from a perfect force source. These limitations include impedance, stiction, and bandwidth. An actuator's impedance is the additional force created at the output by load motion. Impedance is a function of the frequency of the load motion, typically increasing with frequency of load motion. An easily backdriveable system is considered to have low impedance. Stiction describes the phenomenon of stick-slip friction, which is present in most devices where mechanical components are in sliding contact. Stiction must be overcome by a breakaway force, which limits the smallest force the actuator can output. The bandwidth of an actuator is the frequency up to which forces can be accurately commanded. Bandwidth is affected by saturation of power elements, mechanical stiffness, and control system gain among other things. In a perfect force source, impedance is zero (completely back drivable), stiction is zero, and bandwidth is infinite. Muscle has extremely low impedance and stiction and moderate bandwidth and is the currently best known actuation technology that approaches a perfect force source.

Present day actuator technologies have characteristics that have severely limited their use in force-controlled applications. A geared electric motor has a high reflected-inertia, a lot of stiction, and is difficult to back drive. Hydraulic systems have high

- *Arumugom.S is with the Department of Electrical Engg., Udaya School of Engineering, Tamilnadu, India.*
- *Muthuraman.S is with the Dept. of Mechanical Engg., Sun College of Engg. and Tech., Tamilnadu,India.*
- *Ponselvan.V is with the Department of Electrical Engineering, Udaya School of Engineering, Tamilnadu, India.*

seal friction and are often nearly impossible to back drive. By adding Series Elasticity to these conventional systems, a force-controllable actuator with low impedance, low friction, and good bandwidth will result. Below we describe other state-of-the-art methods for force-controllable actuators.

Then we describe Series Elastic Actuators in some detail. Finally, we describe some typical applications for Series Elastic Actuators, highlighting some previous legged robots and exoskeletons that were enabled with Series Elastic Actuators.

## 2. STATE-OF-THE-ARTFORCE CONTROL TECHNOLOGIES

Traditional technologies for force control include current control with direct drive actuation, current control with a geared actuator, current control with low-friction cable drive transmissions, load cells with force feedback, and fluid pressure control. In a direct drive actuator, a high quality servomotor is directly connected to the load and the torque output is accurately controlled using the relation between motor torque and motor current. However, servomotors operate inefficiently at the low speeds and high torques required in most robotic applications. To compensate for their small torques, direct drive servomotors are selected with a power rating much higher than the actual useful power output. This practice results in an ungainly and expensive design. While direct drive actuators are a good approximation of a perfect force source, they are typically too large and heavy for robots that must support the weight of their actuators. Their use is therefore limited to applications where the actuator can be placed in a non-moving base of the robot.

Alternately, smaller and lighter servomotors can be used in low speed/high torque applications if a gear reduction is used to reduce the speed and increase the torque of the motor output. The reduction allows the motor to operate in its sweet spot (high speed/low torque), while providing the low speed/high torque output characteristics desirable in most robotic applications. Current control can then be applied to the geared actuator to control force output. A gear reduction has the major drawbacks of introducing significant friction and of increasing the reflected inertia at the output of the gearbox. In fact, friction can become essentially infinite in some types of non-backdriveable gear reductions with large reduction factors. Such a system would result in extremely poor force fidelity. Since the reflected inertia seen at the output of the actuator increases by the square of the gear ratio, the actuator's impedance also becomes extremely large. Given these performance characteristics, servomotors with gear reductions relying on current control are unsuitable for use in applications requiring high quality force controlled actuators. Significant improvements can be made to geared actuators by using cable drive transmission in lieu of conventional gear reductions. Geared actuators have non-linear, non-continuous dynamics such as stiction and backlash that are hard to model and thus hard to compensate for. Cable drive transmissions, on the other hand, have low stiction and low backlash. Their dynamics are fairly linear and easy to model. Dynamic models of the drive train can then be used in the force controller to help mask the effects of actuator inertia and viscous friction. However, cable drives require large pulleys in order to get a significant transmission ratio. In many applications, space is too limited to accommodate these large pulley systems.

To mitigate the effects of friction and inertia introduced in conventional geared actuators, as described above, a load cell and a feedback control algorithm can be used. The load cell measures the force imparted on the load by the actuator. The feedback controller calculates the error between the measured force and the desired force and applies the appropriate current to the motor to correct any discrepancies. The active force sensing and closed loop control work together to decrease the effects of friction and inertia, thereby attaining a higher force fidelity and lower impedance than with current control alone. However, the load cell method has several shortcomings. First, a stiff load cell can present stability problems.

Consider the case of a stiff load cell between a linear actuator and rigid load. Even a slight linear movement will generate extremely large force readings on the load cell. A high-gain feedback controller would quickly pull the actuator away from the load, causing the force to drop rapidly. The result would be chatter between the actuator and the load. To avoid chatter and maintain stability, the closed loop control gains must be kept very low. The result is a sluggish control system that is unable to respond to small forces. Thus, the effects of friction and inertia cannot be completely masked with the closed loop control system. In addition, shock loads can easily damage the actuator system if a stiff load cell is employed. Both the load cell and gear reduction are susceptible to frequent and expensive damage. While electric actuators can control force by controlling current, pneumatic and hydraulic systems can control force by controlling pressure. In both cases, seal friction can significantly hamper the ability to produce small forces.

Pneumatic systems also suffer from low power density and are difficult to position control. Hydraulic systems generally have high impedance due to both seal friction and large fluidic inertia. Fluidic muscles or McKibben Muscles are pneumatic actuators that deform an elastomeric tube to create a contracting force. While fluidic muscles do not have any sliding seals and therefore can reliably produce small forces, they are generally not a good choice for force control applications due to their nonlinear response, hysteresis, and small stroke to length ratio.

## 3. SERIES ELASTIC ACTUATORS

Series Elastic Actuators have low impedance and friction, and thus can achieve high quality force control. They therefore are well suited for robots in unstructured environments. In Series Elastic Actuators, stiff load cells (which are delicate, expensive, and induce chatter) are replaced with a significantly compliant elastic element (which is robust, inexpensive, and stable). Figure 1 shows the architecture of Series Elastic Actuators. Note that Series Elastic Actuators are topologically similar to any motion actuator with a load sensor and closed loop control system.

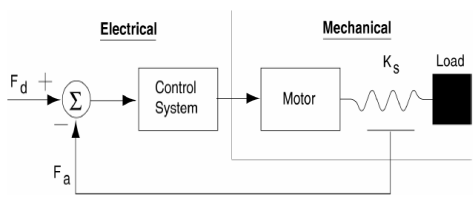

Figure 1: Schematic diagram of a Series Elastic Actuator. A spring is placed between the motor and the load. A control system servos the motor to reduce the difference between the desired force and the measured force signal. The motor can be electrical, hydraulic, pneumatic, or other traditional servo system.

Similar to the load cell method, Series Elastic Actuators use active force sensing and closed loop control to diminish the effects of friction and inertia. By measuring the compression of the compliant element, the force on the load can be calculated using Hooke's Law. A feedback controller calculates the error between the actual force and the desired force, applying appropriate current to the motor to correct any force errors.

In contrast to the load cell method, Series Elasticity introduces significant compliance between the actuator's output and the load, allowing for greatly increased control gains. Consider, as above, the case of a compliant spring between a linear actuator and rigid load. A moderate linear movement will generate a very small force reading. Thus, closed loop control gains can be very high while still insuring the absence of chatter and presence of stability. Increased control gains greatly reduce impedance (increase back-driveability) and reduce the effects of stiction to give the actuators clean force output. Because high impedance and high stiction components are tolerable, the cost and weight can be reduced by allowing the use of smaller, low precision actuator components and replacing expensive load cells with simple springs and position transducers (encoders, potentiometers). These improvements can be realized in both electric and hydraulic actuation domains.

Compared to an actuator with a stiff load cell, Series Elastic Actuators have the following benefits:

1. The actuators exhibit lower output impedance and back-driveability, even in hydraulic systems. The dynamic effects of the motor inertia and gear train friction (or fluidic inertia and seal friction) are nearly invisible at the output. In traditional systems, the actuator dynamics often dominate the mechanism dynamics, making it difficult to accomplish tasks that require high force fidelity.
2. Shock tolerance is greatly improved by the spring placed in series between the drive train and the load.

3. The force transmission fidelity (or smoothness) of the gear reduction or piston is no longer critical, allowing inexpensive gear reduction to be used. Gears typically transmit position with much higher fidelity than force. The series elasticity serves as a transducer between gear reduction output position and load force, greatly increasing the fidelity of force control.

4. The motor's required force fidelity is drastically reduced, allowing inexpensive motors to be used. It is the motor shaft's position, not its output torque that is responsible for the generation of load force. As a result, motors with large torque ripple can be used.

5. Force control stability is improved, even in intermittent contact with hard surfaces. Chatter is eliminated since a relatively large spring deflection is required to exert a small force.

6. Energy can be stored and released in the elastic element, potentially improving efficiency in harmonic applications. Animals commonly utilize the elasticity of tendons to store energy in one part of a locomotive cycle and release it in another, with the muscle doing much less work overall than would otherwise be required. Series Elastic Actuators may allow the same effect to happen in robots, exoskeletons, or other applications, thereby extending their range.

7. The actuator has lower passive impedance at high frequencies. Traditional actuators have impedance that resembles a large inertia (the motor's rotor inertia multiplied by the square of the gear ratio) at high frequencies. A Series Elastic Actuator looks like a spring at high frequencies, which is much more forgiving of collisions and other unexpected interactions.

Robotics, Inc. has developed two commercially available electro-magnetic linear Series Elastic Actuators (Figure 2). These actuators are being used in robot arms, legged robots, exoskeletons, and industrial applications. Due to the simplicity of design, custom Series Elastic Actuators can also be easily developed and retrofitted on legacy systems.

From a macro level, the actuator can be considered to consist of two subassemblies: a drive train subassembly and an output carriage subassembly. The output carriage subassembly is shown in an exploded view apart from the drive train subassembly in Figure 3. When assembled, the output carriage subassembly, a rigid structure, is coupled to the drive train subassembly through the die compression springs. Spring retaining plates firmly sandwich the die springs and ball nut flange. Guide rails pass through bushings in the ball nut flange, forcing the ball nut into linear motion when the ball screw spins. The guide rails also pass through bushings in the spring retaining plates, forcing the entire output carriage subassembly to follow the linear motion of the ball nut.

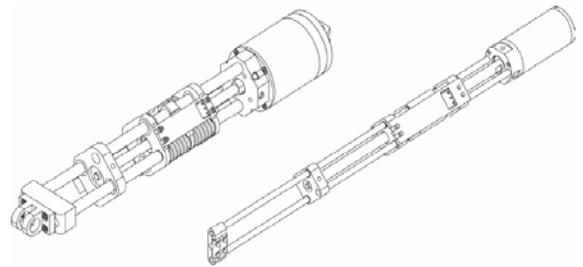

Figure 2: CAD rendering of two electric Series Elastic Actuators.

During operation, the servomotor directly drives the ball screw, converting rotary motion to linear motion of the ball nut. When the motor rotates, the ball nut moves fore or aft on the ball screw depending on the direction of motor rotation. The ball nut flange pushes on two (of four) die compression springs. In turn, the two 'active' die compression springs push on the corresponding spring retaining plate. Spring retaining plates are rigidly attached to output plungers, which are directly connected to the load clevis. Thus, rotary motion of the motor is converted to linear motion of a ball nut which pushes on die compression springs that transmit forces to the load. The force on the load is calculated by measuring the compression of the die springs with a linear potentiometer spanning the spring retaining plates. Figure 4 shows the actuator applying zero force, +300lbs, and -300 lbs. A proportional-derivative (PD) control loop is used to control the actual force on the load.

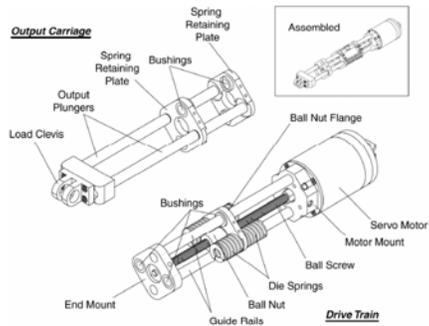

Figure 3: Exploded view of an electric Series Elastic Actuator (version SEA23-23) showing output carriage disassembled from drive train.

We are currently developing hydraulic Series Elastic Actuators for an Army sponsored SBIR to build a gas powered, hydraulically driven quadrupedal robot. These actuators are similar in operation to the electrical Series Elastic Actuators except that a hydraulic cylinder replaces the motor and ball screw. Hydraulics have impressive power to weight compared to electric motors. However, they are often more difficult to use in applications requiring good force control, such as legged robots, due to their poor backdrivability. By implementing Series Elastic Actuators with hydraulic cylinders, the result is an excellent force generator. Figure 5 shows an exploded view of a hydraulic Series Elastic Actuator that can produce 1300 lbs of force and achieve a force-control bandwidth of 50 Hz.

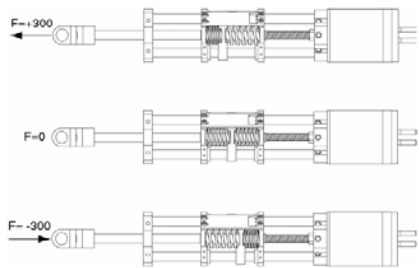

Figure 4: Actuator shown in unloaded, +300lb load, and -300lb load configurations.

If either the electric or the hydraulic actuators employed a rigid load cell in place of the series elastic element, they would have high output impedance and high friction, and thus be poor force sources. However, with Series Elasticity, the actuators are very force sensitive and can achieve high bandwidth at moderate force amplitudes and have low output impedance. The lowest resolvable force, which is limited by residual stiction, is approximately one pound.

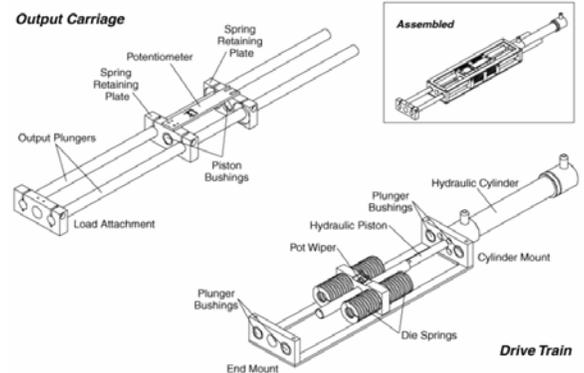

Figure 5: Exploded view of a hydraulic Series Elastic Actuator (version HyEA75-32) showing output carriage disassembled from drive train.

In Table 1 the Series Elastic Actuators are compared with traditional actuation methods. Series Elasticity improves the force fidelity of gear motors and hydraulics so they are comparable to direct drive motors without sacrificing high force/torque capabilities. They are lighter than direct drive motors in high force applications since they can use an optimal gear reduction. Position controllability remains good as a position servo can be implemented on top of the force control servo. Some bandwidth is lost in that process. However, we have found that the achievable position control bandwidth of Series Elastic Actuators is still higher than that of muscle.The specification of each actuator can be seen below in Table 2.

Table 1: A comparison of major actuation technologies.

| Actuation Type | Max. Force | Max. Speed | Low Force Ability | Position Controllability | Back-drivability |
|---|---|---|---|---|---|
| Pneumatic | Med. | Med. | Fair, Stiction | Poor | Fair |
| Hydraulic | High | Med. | Poor, Stiction | Good | Poor |
| Direct drive electric | Low | High | Excellent | Good | Excellent |
| Electric gear motor | Med./High | High. | Poor, Friction | Good | Poor |
| Electric Series Elastic Actuator | Med/High | High | Excellent | Good | Excellent |
| Hydraulic Series Elastic Actuator | High | Med. | Excellent | Good | Excellent |

Table 2: Specifications of four Series Elastic Actuators. Versions SEA-23-23 and SEA-12-25 have brushless DC motors. Versions HyEA-75-32 and HyEA-50-31 have hydraulic cylinders. The weight measurement does not include motor amplifiers or energy source for the electric actuators nor hydraulic pumps, accumulators, or energy sources for the hydraulic actuators. For the hydraulic actuators, the continuous and intermittent specifications are identical as we assume that the hydraulic power circuit can provide adequate power and fluid cooling. In many applications, particularly legged robots, the power limit will be due to those components.

| | SEA-23-23 | SEA-12-25 | HyEA-75-32 | HyEA-50-31 |
|---|---|---|---|---|
| Weight | 2.5 lbs | 1 lb | 6 lbs | 5.5 lbs |
| Stroke | up to 12 in. | up to 12 in. | up to 12 in. | up to 12 in. |
| Diameter | 2.3 in | 1.2 in | *** | *** |
| Length | stroke + 6 in. | stroke + 6 in. | 2 x stroke + 4 in. | 2 x stroke + 4 in. |
| Maximum Speed | 11 in/s | 1.9 in/s | 122 in/s | 132 in/s |
| Continuous Force | 127 lbs | 30 lbs | 1324 lbs | 588 lbs |
| Continuous Power | 0.22 hp | 0.06 hp | 24.5 hp | 11.8 hp |
| Intermittent Force | 300 lbs | 87 lbs | *** | *** |
| Intermittent Power | 0.85 hp | 0.174 hp | *** | *** |
| Continuous Power to Weight | 0.088 hp/lb | 0.06 hp/lb | 4.1 hp/lb | 2.1 hp/lb |
| Intermittent Power to Weight | 0.34 hp/lb | 0.174 hp/lb | *** | *** |
| Small Force Bandwidth | 35 Hz | 25 Hz | 50 Hz | 50 Hz |
| Large Force Bandwidth | 7.5 Hz | 5 Hz | 10 Hz | 10 Hz |
| Operating Voltage/Pressure | 24-48 Volts | 24-48 Volts | 3000 psi | 3000 psi |
| Maximum Current/Flow | 20 amps | 10 amps | 54 cis | 26 cis |

# 4. APPLICATIONS

## 4.1. Legged Robots

Actuators with muscle-like properties could allow legged robots to achieve performance approaching that of their biological counterparts. Some of the beneficial properties of muscle include its low impedance, high force-fidelity, low friction, and good bandwidth. Series Elastic Actuators share these beneficial properties with muscle and thus are well suited for legged robots. These high quality force-controllable actuators allow the control system to exploit the natural dynamics of the robot, to distribute forces among the legs, and to provide an active suspension that is robust to rough terrain.

Most airplanes are designed to have wings so that they glide stably, requiring only a simple power source and simple control to fly; early locomotives used fly-ball governors, a mechanical feedback device, to help maintain constant speed; satellites and rifle bullets spin to stabilize their trajectory. These machines were designed so that their natural dynamics allow minimal control effort. Animals have evolved similar mechanisms that exploit natural dynamics. Birds have wings that allow for stable gliding. Fish have hydro-dynamically stable shapes and are neutrally buoyant.

Running animals have springy legs. And walking animals have joint limits, compliant ankles, and legs that swing passively. Similarly, natural dynamics can be exploited in the control of bipedal walking robots: the swing-leg can swing freely once started; a kneecap can be used to prevent the leg from inverting; and a compliant ankle can be used to naturally transfer the center of pressure along the foot and help in toe off. Each of these mechanisms helps make control easier to achieve and results in motion that is smooth and natural looking.

In order to exploit passive dynamics in a robot, the actuators must present extremely low impedance and friction to the system. Unfortunately, with traditional actuation systems such as hydraulics and highly geared motors, the output impedance and inertia are high. In contrast, Series Elastic Actuators present extremely low impedance and low friction and thus may be used in robots that exploit their natural dynamics. Like a multi-legged table, a multi-legged robot may present an indeterminate, over constrained system with regards to the ground reaction forces on each foot. One of the legs of the table may be off the ground, allowing for rocking of the table. Similarly, in a multi-legged robot that uses rigid position control, the legs may produce conflicting forces on the body, causing foot slippage and large internal forces. Series Elastic Actuators can prevent these problems by providing for accurately distributed forces among the feet of the robot, which can be independent of the position of the feet. The high force fidelity of a Series Elastic Actuator provides for stability and prevents slippage, giving the robot a sure footing.

Series Elastic Actuators provide for an effective active suspension system due to their high-force fidelity and low impedance. Active suspensions can be used to arbitrarily orient and isolate payloads and can impart restoring forces to a robot's center of mass, allowing for dynamic balancing. These active suspension behaviors are not possible when operating a purely position-controlled robot in an unknown environment. For this reason, robots that follow rigid trajectories have trouble walking over rough terrain because they cannot conform to the terrain or apply restoring forces dynamically to the robot. In contrast, Series Elastic Actuators may be used to control the net interaction forces between the robot and the ground, providing an active suspension system independent of the position of the feet or the roughness of the terrain.

M2 and Spring Flamingo (Figure 6), two bipedal walking robot developed at the MIT Artificial Intelligence Lab by Yobotics co-founders, used Series Elastic Actuators to push the envelope of walking robotics. M2 has 12 active degrees of freedom and Spring Flamingo has 6, each employing an electric Series Elastic Actuator.

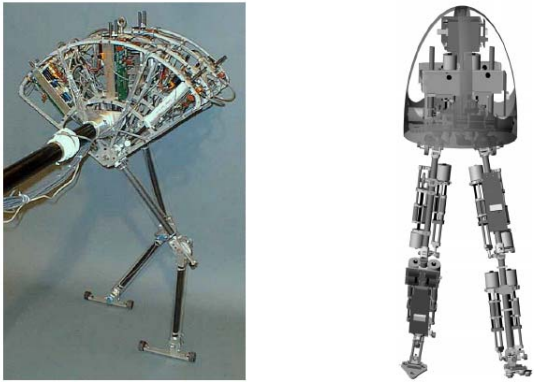

Figure 6: Spring Flamingo (left) and M2 (CAD rendering at right), two bipedal walking robots that employ Series Elastic Actuators. Spring Flamingo is a planar, 6 degree-of-freedom biped, while M2 is a 3D, 12 degree-of-freedom biped.

## 4.2. Performance Amplifying Exoskeletons and Haptic Interfaces

Series Elastic Actuators, with their low impedance and high force fidelity, are ideal for exoskeletons, haptic interfaces, and other wearable robots. In these applications, the device should not inhibit the natural fluidic motions of the user. An exoskeleton should supply assistive forces to the joints during power-intensive activities (such as knee extension when climbing stairs, or carrying heavy loads) without inhibiting motion of the user during low-power activities (such as leg swing and foot placement). A haptic interface should provide force feedback when touching virtual surfaces or performing physical actions, but should provide no impedance to motion when moving in free space. Ideally, these devices would be perfectly transparent, creating the feeling that the device is not even there.

Series Elastic Actuators are a good choice for approximating perfect transparency in these devices because of their low impedance. Yobotics has used Series Elastic Actuators in the development of a powered exoskeleton (Figure 7) that improves strength and endurance of the operator.

### 4.3. Robotic Arms

When robotic arms must operate in uncertain environments, especially when interacting with people, Series Elastic force control can improve performance and safety. By imposing maximum force limits in the joints, operators can work within the workspace of a powerful robot with an added measure of safety. If an operator contacts any point on the robot, a fault will be registered very quickly. Traditional robots would require excessive load sensors throughout the arm to operate safely in an unstructured environment. The inherent shock tolerance of Series Elastic Actuators also allows the robot to crash with minimal or no damage.

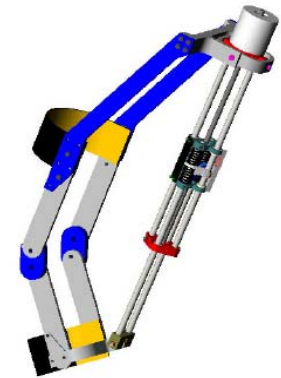

Figure 7: The RoboKnee, an exoskeleton for enhancing human strength and endurance. CAD rendering (left) shows the knee brace and electric Series Elastic Actuator.

Performance can be greatly improved using force control in situations where it is better for the physical constraints of the environment to dictate the motion of the robot as opposed to a preprogrammed trajectory. Peg-in-hole type problems are a classic example of this. Instead of precisely locating a work piece or using a vision system feedback to mate two parts, a robot with Series Elastic Actuators could do what people do and simply push down and randomly move the peg around in the vicinity of the hole until it drops in. Series Elasticity also provides an advantage in assembly operations that require a certain insertion force or torque. Instead of using specially instrumented endeffectors for every operation the robot can use standard tools.

Yobotics has developed a 6 degree-of-freedom force controlled robotic arm for the MIT Media Lab for use in computer learning research. The key functionality was the ability for an operator to grab the robot while it is moving. The robot can sense the interaction of the operator and go into an 'anti-gravity' mode in which the actuators compensate for the weight of the arm allowing the operator to easily teach the robot a new trajectory. A CAD model of the robot arm and gripper can be seen in Figure 8.

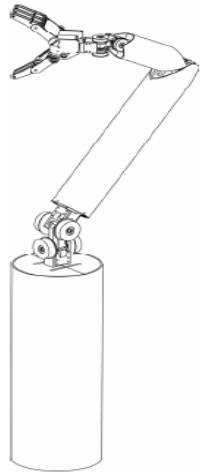

Figure 8: CAD model of Agile Robotic Arm with 6 force controlled degrees of freedom.

## 5  CONCLUSIONS

Series Elastic Actuators' ability to closely approximate a pure force source allows them to better integrate into many robotic applications, including legged robots, human-in-the-loop systems, haptic interfaces, and robotic arms. This technology is independent of actuation method and has been proven in both electric and hydraulic applications. This cross-platform suitability allows Series Elastic Actuator technology to be readily integrated into new machines as well as retrofitted into legacy equipment.

Mr.S.Arumugom M.E., Electrical Engineering Degree holder, He has passed M.E. Power systems in first class with distinction. He has teaching and industrial experience of about nine years. He is currently working as Asst. Professor in Udaya School of Engineering, Tamilnadu, India.

Mr.S.Muthuraman is working as Assistant Professor in Dept. of Mechanical Engineering for Sun College of Engineering and Technology. He has passed M.E with First class Distinction. He has Teaching and Industrial Experience of about nine Years.

Mr.V.Ponselvan M.E., Electrical Engineering Degree holder, He has passed M.E. Applied Electronics in first class. He has teaching and industrial experience of about three years. He is currently working as Lecturer in Udaya School of Engineering.